\documentclass[11pt,a4paper]{article}
\usepackage[hyperref]{acl2019}
\usepackage{times}
\usepackage{latexsym}

\usepackage{url}
\usepackage{amsmath,amssymb}
\usepackage{graphicx}
\usepackage{bm}
\usepackage{textcomp}
\usepackage{multirow}
\usepackage{color}
\usepackage{url}
\usepackage{subcaption}
\usepackage{booktabs}
\usepackage{soul}
\graphicspath{{figures/}}

\def\figref#1{Fig.~\ref{#1}}
\def\secref#1{Sec.~\ref{#1}}
\def\tabref#1{Table~\ref{#1}}
\def\eqnref#1{Eqn.~\ref{#1}}

\aclfinalcopy 

\newcommand\numberthis{\addtocounter{equation}{1}\tag{\theequation}}

\title{Explore, Propose, and Assemble: \\ An Interpretable Model for Multi-Hop Reading Comprehension}
\author{Yichen Jiang\thanks{\:\:equal contribution; part of this work was done during the second author's internship at UNC (from IIT Bombay).} \;\;\;\;\;\; Nitish Joshi\footnotemark[1] \;\;\;\;\;\; Yen-Chun Chen \;\;\;\;\;\; Mohit Bansal \\
  UNC Chapel Hill \\
  {\tt \{yichenj, nitish, yenchun, mbansal\}@cs.unc.edu} \\
 }
 
\date{}

\begin{document}
\maketitle
\begin{abstract}
Multi-hop reading comprehension requires the model to explore and connect relevant information from multiple sentences/documents in order to answer the question about the context. 
To achieve this, we propose an interpretable 3-module system called Explore-Propose-Assemble reader (EPAr). 
First, the Document Explorer iteratively selects relevant documents and represents divergent reasoning chains in a tree structure so as to allow assimilating information from all chains.
The Answer Proposer then proposes an answer from every root-to-leaf path in the reasoning tree. 
Finally, the Evidence Assembler extracts a key sentence containing the proposed answer from every path and combines them to predict the final answer.
Intuitively, EPAr approximates the coarse-to-fine-grained comprehension behavior of human readers when facing multiple long documents.
We jointly optimize our 3 modules by minimizing the sum of losses from each stage conditioned on the previous stage's output. 
On two multi-hop reading comprehension datasets WikiHop and MedHop, our EPAr model achieves significant improvements over the baseline and competitive results compared to the state-of-the-art model.
We also present multiple reasoning-chain-recovery tests and ablation studies to demonstrate our system's ability to perform interpretable and accurate reasoning.\footnote{Our code is publicly available at: \\ \url{https://github.com/jiangycTarheel/EPAr}}

\end{abstract}  
\section{Introduction}
\label{sec:intro}
\begin{figure*}[t]
\centering
\begin{subfigure}{0.495\textwidth}
\includegraphics[width=0.98\linewidth]{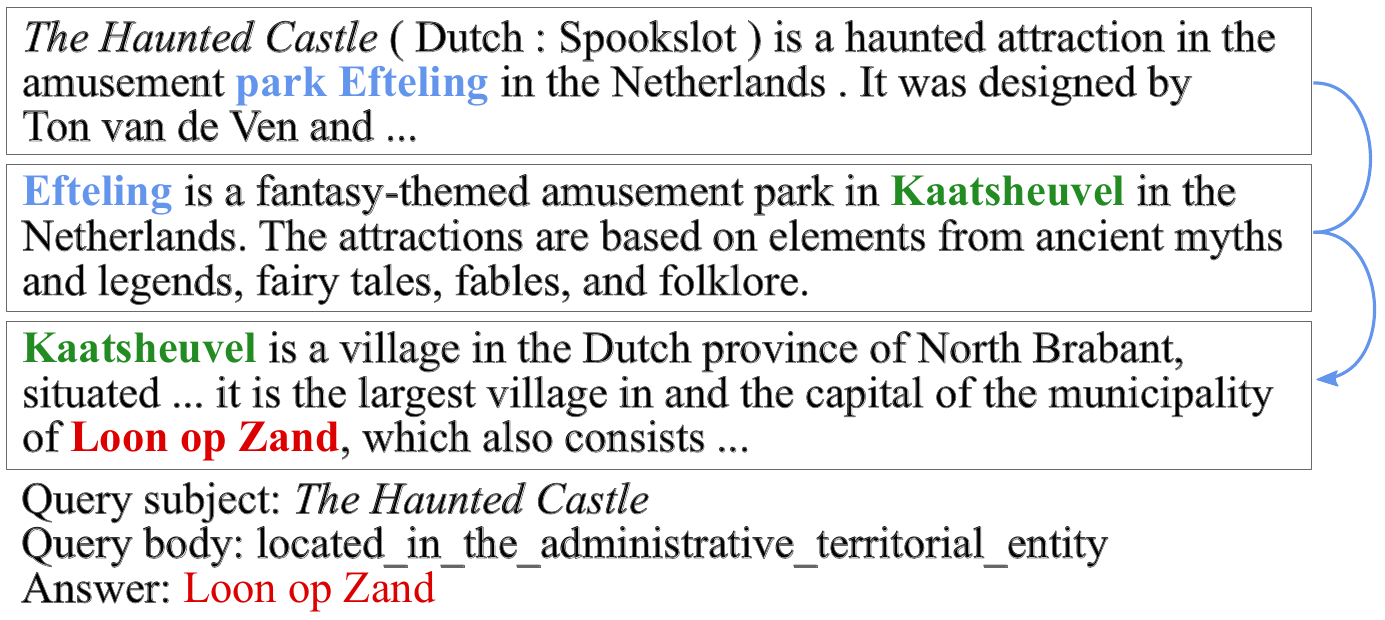} 
\caption{}
\label{fig:example-intro-a}
\end{subfigure}
\begin{subfigure}{0.495\textwidth}
\includegraphics[width=0.98\linewidth]{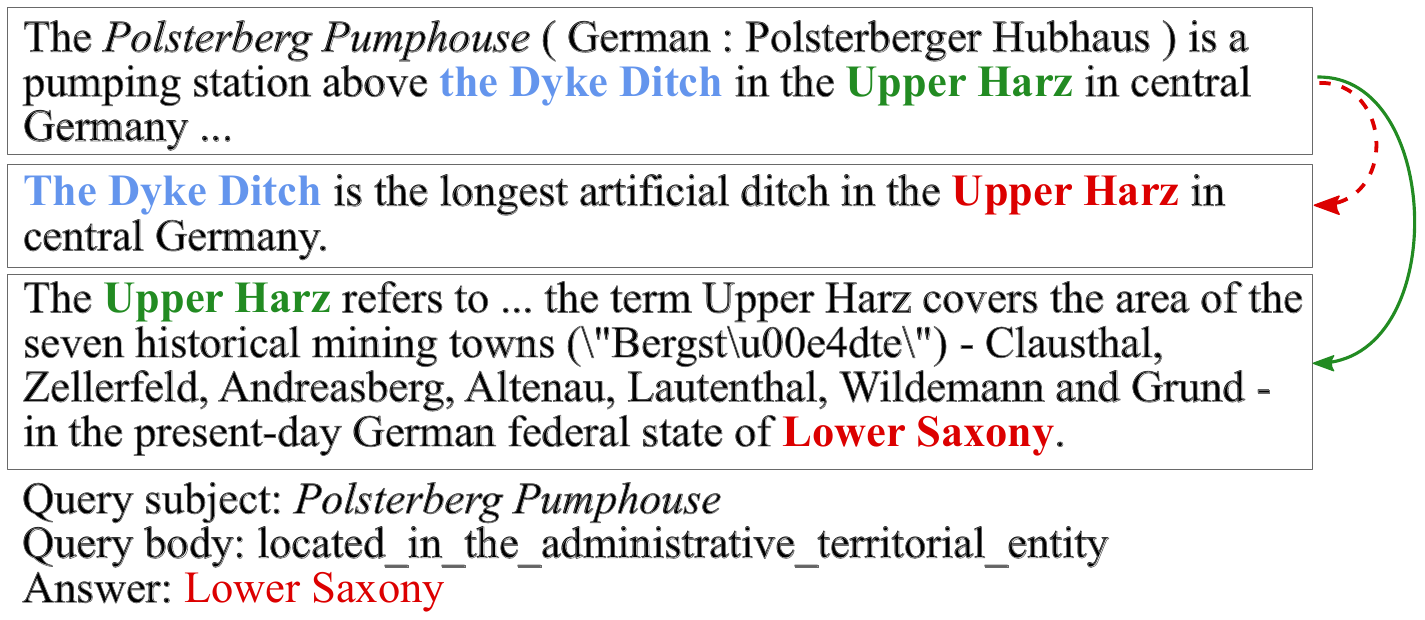}
\caption{}
\label{fig:example-intro-b}
\end{subfigure}
\vspace{-7pt}
\caption{Two examples from the QAngaroo WikiHop dataset where it is necessary to combine information spread across multiple
documents to infer the correct answer. \textbf{(a)}: The hidden reasoning chain of 3 out of a total of 37 documents for a single query. \textbf{(b)}: Two possible reasoning chains that lead to different answers: ``Upper Harz'' and ``Lower Saxony'', while the latter (green solid arrow) fits better with query body ``administrative territorial entity''. }
\vspace{-7pt} 
\end{figure*}
The task of machine reading comprehension and question answering (MRC-QA)
requires the model to answer a natural language question by finding relevant information and knowledge in a given natural language context.
Most MRC datasets require single-hop reasoning only, which means that the evidence necessary to answer the question is concentrated in a single sentence or located closely in a single paragraph.
Such datasets emphasize the role of locating, matching, and aligning information between the question and the context.
However, some recent multi-document, multi-hop reading comprehension datasets, such as WikiHop and MedHop~\citep{welbl2017qangaroo}, have been proposed to further assess MRC systems' ability to perform multi-hop reasoning, where the required evidence is scattered in a set of supporting documents.

These multi-hop tasks are much more challenging than previous single-hop MRC tasks~\citep{rajpurkar2016squad,rajpurkar2018squad2,Hermann:cnndm,nguyen2016msmarco,yang2015wikiqa} for three primary reasons.
First, the given context contains a large number of documents (e.g., 14 on average, 64 maximum for WikiHop). Most existing QA models cannot scale to the context of such length, and it is challenging to retrieve a reasoning chain of documents with complete information required to connect the question to the answer in a logical way.
Second, given a reasoning chain of documents, it is still necessary for the model to consider evidence loosely distributed in all these documents in order to predict the final answer.
Third, there could be more than one logical way to connect the scattered evidence (i.e., more than one possible reasoning chain) and hence this requires models to assemble and weigh information collected from every reasoning chain before making a unified prediction.

To overcome the three difficulties elaborated above, we develop our interpretable 3-module system based on examining how a human reader would approach a question, as shown in \figref{fig:example-intro-a} and \figref{fig:example-intro-b}. 
For the 1st example, instead of reading the entire set of supporting documents sequentially, she would start from the document that is directly related to the query subject (e.g., ``The Haunted Castle").
She could then read the second and third document by following the connecting entities ``park Efteling" and ``Kaatsheuvel", and uncover the answer ``Loon op Zand" by comparing phrases in the final document to the query.
In this way, the reader accumulates knowledge about the query subject by exploring inter-connected documents, and eventually uncovers the entire reasoning chain that leads to the answer.
Drawing inspiration from this coarse (document-level) plus fine-grained (word-level) comprehension behavior, we first construct a $T$-hop Document Explorer model, a hierarchical memory network, which at each recurrent hop, selects one document to read, updates the memory cell, and iteratively selects the next related document, overall constructing a reasoning chain of the most relevant documents. 
We next introduce an Answer Proposer that performs query-context reasoning at the word-level on the retrieved chain and predicts an answer. Specifically, it encodes the leaf document of the reasoning chain while attending to its ancestral documents, and outputs ancestor-aware word representations for this leaf document, which are compared to the query to propose a candidate answer.

However, these two components above cannot handle questions that allow multiple possible reasoning chains that lead to different answers, as shown in \figref{fig:example-intro-b}. After the Document Explorer selects the 1st document, it finds that both the 2nd and 3rd documents are connected to the 1st document via entities ``the Dyke Ditch" and ``Upper Harz" respectively.
This is a situation where a single reasoning chain diverges into multiple paths, and it is impossible to tell which path will lead to the correct answer before finishing exploring all possible reasoning chains/paths.
Hence, to be able to weigh and combine information from multiple reasoning branches, the Document Explorer is rolled out multiple times to represent all the divergent reasoning chains in a `reasoning tree' structure, so as to allow our third component, the Evidence Assembler, to assimilate important evidence identified in every reasoning chain of the tree to make one final, unified prediction.
To do so, the Assembler selects key sentences from each root-to-leaf document path in the `reasoning tree' and forms a new condensed, salient context which is then bidirectionally-matched with the query representation to output the final prediction. 
Via this procedure, evidence that was originally scattered widely across several documents is now collected concentratedly, hence transforming the task to a scenario where previous standard phrase-matching style QA models~\citep{seo2016bidaf,xiong2016dynamic,dhingra2017GA} can be effective.

Overall, our 3-module, multi-hop, reasoning-tree based EPAr (Explore-Propose-Assemble reader) closely mimics the coarse-to-fine-grained reading and reasoning behavior of human readers.
We jointly optimize this 3-module system by having the following component working on the outputs from the previous component and minimizing the sum of the losses from all 3 modules.
The Answer Proposer and Evidence Assembler are trained with maximum likelihood using ground-truth answers as labels, while the Document Explorer is weakly supervised by heuristic reasoning chains constructed via TF-IDF and documents with the ground-truth answer. 

\begin{figure*}[t]
\centering
\includegraphics[width=0.98\textwidth]{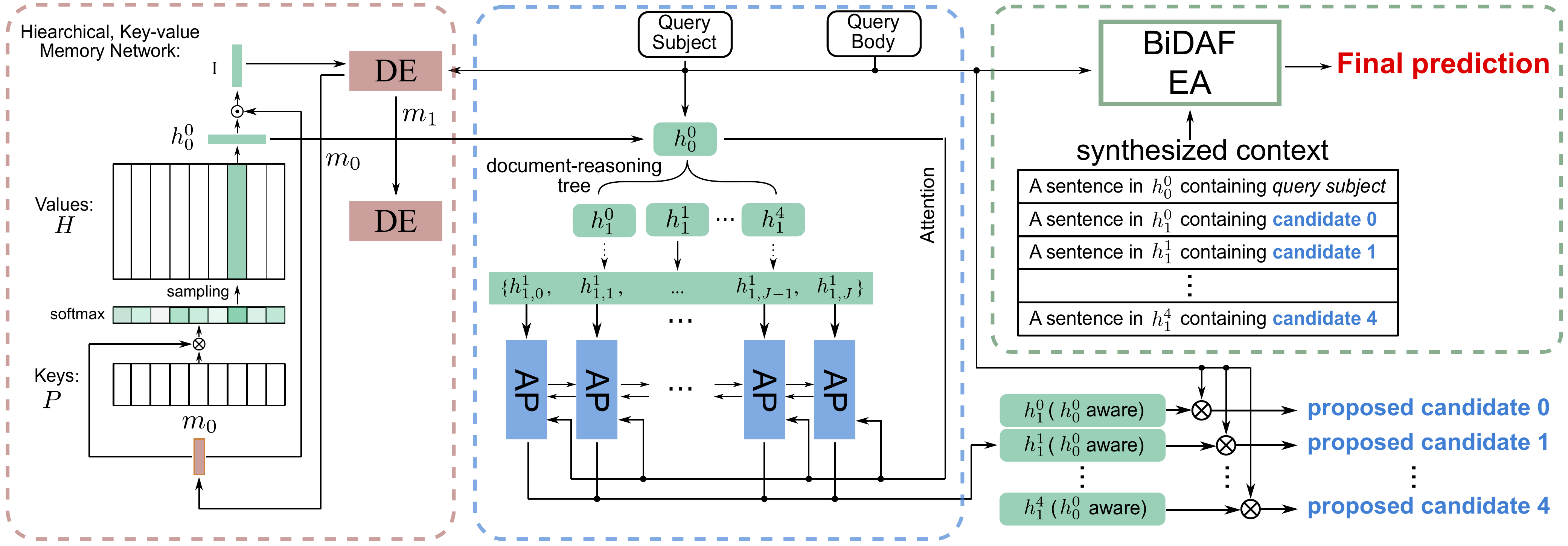}
\vspace{-4pt}
\caption{The full architecture of our 3-module system EPAr, with the Document Explorer (DE, left), Answer Proposer (AP, middle), and Evidence Assembler (EA, right). \label{fig:model}
\vspace{-7pt}
}
\end{figure*}

On WikiHop, our system achieves the highest-reported dev set result of 67.2\%, outperforming all published models\footnote{At the time of submission: March 3rd, 2019.} on this task, and 69.1\% accuracy on the hidden test set, which is competitive with the current leaderboard state-of-the-art. 
On MedHop, our system outperforms all previous models, achieving the new state-of-the-art test leaderboard accuracy. 
It also obtains statistically significant (p \(< 0.01\)) improvement over our strong baseline on the two datasets.
Further, we show that our Document Explorer combined with 2-hop TF-IDF retrieval is substantially better than two TF-IDF-based retrieval baselines in multiple reasoning-chain recovery tests including on human-annotated golden reasoning chains.
Next, we conduct ablations to prove the effectiveness of the Answer Proposer and Evidence Assembler in comparison with several baseline counterparts, and illustrate output examples of our 3-module system's reasoning tree.

\section{Model}
\label{sec:model}

In this section, we describe our 3-module system that constructs the `reasoning tree' of documents and predicts the answer for the query.
Formally, given a query $q$ and a corresponding set of supporting documents $D = \{d_i\}_{i=1}^N$, our system tries to find a reasoning chain of documents $d_{1}^{\prime}, \dots, d_{T}^{\prime}, d_{i}^{\prime} \in D$.\footnote{In WikiHop dataset, $T \leq 3$.} 
The information from these selected documents is then combined to predict the answer among the given answer candidates.
In the WikiHop and MedHop datasets, a query consists of a subject $q_{sub}$ (e.g., ``The Haunted Castle" in \figref{fig:example-intro-a}) and a body $q_{bod}$ (e.g., ``located in the administrative territorial entity").
There is one single correct answer $a$ (e.g., ``Loon op Zand") in the set of candidate answers $A = \{c_l\}_{l=1}^L$ such that the relation $q_{bod}$ holds true between $q_{sub}$ and $a$.

\subsection{Retrieval and Encoding}
\label{ssec:ret_and_enc}
In this section, we describe the pre-processing document retrieval and encoding steps before introducing our three modules of EPAr. We adopt a 2-hop document retrieval procedure to reduce the number of supporting documents that are fed to our system. 
We first select one document with the shortest TF-IDF distance to the query.
We then rank the remaining documents according to their TF-IDF distances to the first selected document and add the top \(N^{\prime}-1\) documents to form the context with a total of $N'$ documents for this query.
Adding this preprocessing step is not only helpful in reducing GPU memory consumption but also helps bootstrap the training by reducing the search space of the Document Explorer (\secref{sec:docexp}).

We then use a Highway Network~\citep{srivastava2015highway} of dimension $d$, which merges the character embedding and GloVe word embedding~\citep{pennington2014glove}, to get the word representations for the supporting documents and query\footnote{Unlike previous works~\citep{welbl2017qangaroo,dhingra2018corefgru,Cao2018question,song2018exploring} that concatenate supporting documents together to form a large context, we instead maintain the document-level hierarchy and encode each document separately.}.
This gives three matrices: \(\mathbf{X} \in \mathbb{R}^{N^{\prime} \times K \times d}\), \(\mathbf{Q_{sub}} \in \mathbb{R}^{J_s \times d}\) and \(\mathbf{Q_{bod}} \in \mathbb{R}^{J_b \times d}\), \(K\), \(J_s\), \(J_b\) are the lengths of supporting documents, query body, and query subject respectively.
We then apply a bi-directional LSTM-RNN~\citep{hochreiter1997lstm} of $v$ hidden units to get the contextual word representations for the documents \(\mathbf{H} = \{h_1, \cdots, h_{N^{\prime}}\}\) s.t. \(h_i \in \mathbb{R}^{K \times 2v}\) and the query \(\mathbf{U_{sub}} \in \mathbb{R}^{J_s \times 2v}\), \(\mathbf{U_{bod}} \in \mathbb{R}^{J_{b} \times 2v}\). 
Other than the word-level encoding, we also collect compact representations of all the supporting documents, denoted as $\mathbf{P} = \{p_1, \cdots, p_{N^{\prime}}\}$, by applying the self-attention mechanism in \citet{zhong2019coarse} (see details in appendix). We obtain embeddings for each candidate \(c_i \in \{c_1,c_2,..,c_L\}\) using the average-over-word embeddings of the first mention\footnote{We tried different approaches to make use of all mentions of every candidate, but observe no gain in final performance.} of the candidate in \(\mathbf{H}\). 

\subsection{Document Explorer}
\label{sec:docexp}
Our Document Explorer (DE, shown in the left part of \figref{fig:model}) is a hierarchical memory network~\citep{chandar2016hierarchical}.
It utilizes the reduced document representations \(\mathbf{P} = \{p_1, p_2, \cdots, p_{N^{\prime}}\}\) and their corresponding word-level representations \(\mathbf{H} = \{h_1, h_2, \cdots, h_{N^{\prime}}\}\) as the key-value knowledge base and maintains a memory \(m\) using a Gated Recurrent Unit (GRU)~\citep{cho2014gru}. At every step, the DE selects a document which is related to the current memory state and updates the internal memory. This iterative procedure thus constructs a reasoning chain of documents.

\paragraph{Read Unit}
At each hop $t$, the model computes a \emph{document-selection distribution} $P$ over every document based on the bilinear-similarity between the memory state \(m\) and document representations \(\mathbf{P}\) using the following equations\footnote{We initialize the memory state with the last state of the query subject \(\mathbf{U_{sub}}\) to make first selected document directly conditioned on the query subject.}:
\vspace{-6pt}
\begin{equation*} \label{eq:docexp-read}
\begin{split}
x_n  =  p_n^T \mathbf{W_r} m^t
\quad
\chi  = \mathrm{softmax}(x)
\quad
P(d_i) = \chi_i
\end{split}
\end{equation*}
The read unit looks at all document (representation) $\mathbf{P}$ and selects (samples) a document $d_i \sim P$.
The write operation updates the internal state (memory) using this sampled document.

\paragraph{Write Unit}
After the model selects \(d_i \in D\),  
the model then computes a distribution over every word in document \(d_i\) based on the similarity between the memory state \(\mathbf{m}\) and its word representations \(h_i \in \mathbf{H}\).
This distribution is then used to compute the weighted average of all word representations in document \(d_i\).
We then feed this weighted average \(\Tilde{h}\) as the input to the GRU cell and update its memory state \(\mathbf{m}\) (subscript $i$ is omitted for simplicity):
\begin{equation} \label{eq:docexp-write}
\begin{split}
w_k & =  h_k^T \mathbf{W_w} m \\
\omega & = \mathrm{softmax}(w) \\
\end{split}
\quad
\begin{split}
\Tilde{h} & = \sum\nolimits_{k=1}^K h_k \omega_k \\
m^{t+1} & = \mathbf{GRU}(\Tilde{h}, m^t)
\end{split}
\vspace{-10pt}
\end{equation}
Combining the `read' and `write' operations described above, we define a recurrent function: \((\hat{h}_{t+1},\ m^{t+1}) = f_{DE}(m^t)\) such that \(\hat{h}_{t+1} \in \mathbf{H}\) and \(\hat{h}_t \neq \hat{h}_{t+1}\).
Therefore, unrolling the Document Explorer for \(T\) hops results in a sequence of \emph{non-repeating} documents \(\mathbf{\hat{H}} = \{\hat{h}_1, \cdots, \hat{h}_T\}\) such that each document \(\hat{h}_i\) is selected iteratively based on the current memory state building up one reasoning chain of documents. 
In practice, we roll out DE multiple times to obtain a document-search `reasoning tree', where each root-to-leaf path corresponds to a query-to-answer reasoning chain.

\subsection{Answer Proposer}
The Answer Proposer (AP, shown in the middle part of \figref{sec:model}) takes as input a single chain of documents \(\{\hat{h}_1, \cdots, \hat{h}_T\}\) from one of the chains in the `reasoning tree` created by the DE, and tries to predict a candidate answer from the last document \(\hat{h}_T\) in that reasoning chain. 
Specifically, we adopt an LSTM-RNN with an attention mechanism~\citep{Bahdanau:14} to encode the \(\hat{h}_T\) to ancestor-aware representations \(y\) by attending to \([\hat{h}_{1, \dots, T-1}]\).
The model then computes a distribution over words \(\hat{h}_T^i\ \in \hat{h}_T\) based on the similarity between \(y\) and the query representation.
This distribution is then used to compute the weighted average of word representations \(\{h_T^1, h_T^2, \cdots, h_T^K\}\).
Finally, AP proposes an answer among all candidates \(\{c_1, \cdots, c_L\}\) that has the largest similarity score with this weighted average \(\Tilde{h}_T\).
\begin{align*}
e_{i}^{k} & = \mathbf{v}^T \mathrm{tanh}(\mathbf{W_h} \hat{h}_{cct}^i + \mathbf{W_s} s^k + \mathbf{b}) \\
a^k & = \mathrm{softmax}(e^k); \;\;\; c^k = \sum\nolimits_{i} a_{i}^{k} h_{cct}^{i}\\
y^{k} &= \mathbf{LSTM}(\hat{h}^{k-1}_T, s^{k-1}, c^{k-1}) \\
w^k &= \boldsymbol{\alpha}(y^{k},u_{s}) + \boldsymbol{\alpha}(y^{k},u_{b}); \;\;\epsilon = \mathrm{softmax}(w) \\
a &= \sum\nolimits_{k=1}^K \hat{h}^k_T \epsilon_k \numberthis; \quad Score_l = \boldsymbol{\beta}(c_l, a)\label{eq:ansprop}
\end{align*}
where $\hat{h}_{cct} = [\hat{h}_{1, \dots, T-1}]$ is the concatenation of documents in the word dimension; \(u_{s}\) and \(u_{b}\) are the final states of \(\mathbf{U_{sub}}\) and \(\mathbf{U_{bod}}\) respectively, and $s^k$ is the LSTM's hidden states at the kth step. The Answer Proposer proposes the candidate with the highest score among \(\{c_1, \cdots, c_L\}\). All computations in ~\eqnref{eq:ansprop} that involve trainable parameters are marked in bold.\footnote{See appendix for the definition of the similarity functions \(\boldsymbol{\alpha}\) and \(\boldsymbol{\beta}\).}
This procedure produces ancestor-aware word representations that encode the interactions between the leaf document and ancestral document, and hence models the multi-hop, cross-document reasoning behavior.

\subsection{Evidence Assembler}

\label{ssec:assembler}
As shown in \figref{fig:example-intro-b}, it is possible that a reasoning path could diverge into multiple branches, where each branch represents a unique, logical way of retrieving inter-connected documents.
Intuitively, it is very difficult for the model to predict which path to take without looking ahead.
To solve this, our system first explores multiple reasoning chains by rolling out the Document Explorer multiple times to construct a `reasoning tree' of documents, and then aggregates information from multiple reasoning chains using a Evidence Assembler (EA, shown in the right part of \figref{sec:model}), to predict the final answer. 
For each reasoning chain, the Assembler first selects one sentence that contains the candidate answer proposed by the Answer Proposer and concatenates all these sentences into a new document \(h^\prime\). This constructs a highly informative and condensed context, at which point previous phrase-matching style QA models can work effectively. 
Our EA uses a bidirectional attention flow model~\citep{seo2016bidaf} to get a distribution over every word in \(h^\prime\) and compute the weighted average of word representations \(\{h^{\prime1}, \cdots, h^{\prime K}\}\) as \(\Tilde{h}^{\prime}\). Finally, the EA selects the candidate answer of the highest similarity score w.r.t. \(\Tilde{h}^{\prime}\).

\subsection{Joint Optimization} 
Finally, we jointly optimize the entire model using the cross-entropy losses from our Document Explorer, Answer Proposer, and Evidence Assembler.
Since the Document Explorer samples documents from a distribution, we use weak supervision at the first and the final hops to account for the otherwise non-differentiabilty in the case of end-to-end training. Specifically, we use the document having the shortest TF-IDF distance w.r.t. the query subject to supervise the first hop and the documents which contain at least one mention of the answer to supervise the last hop. This allows the Document Explorer to learn the chain of documents leading to the document containing the answer from the document most relevant to the query subject.
Since there can be multiple documents containing the answer, we randomly sample a document as the label at the last hop.
For the Answer Proposer and Evidence Assembler, we use cross-entropy loss from the answer selection process.

\section{Experiments and Results}
\subsection{Datasets and Metrics}
\label{ssec:dataset}
We evaluate our 3-module system on the WikiHop and the smaller MedHop multi-hop datasets from QAngaroo~\citep{welbl2017qangaroo}. 
For the WikiHop dev set, each instance is also annotated as ``follows" or ``not follows", i.e., whether the answer can be inferred from the given set of supporting documents, and ``single" or ``multiple", indicating whether the complete reasoning chain comprises of single or multiple documents. 
This allows us to evaluate our system on less noisy data and to investigate its strength in queries requiring different levels of multi-hop reasoning. Please see appendix for dataset and metric details.

\subsection{Implementation Details}
For WikiHop experiments, we use 300-d GloVe word embeddings ~\citep{pennington2014glove} for our main full-size `EPAr' model and 100-d GloVE word embeddings for our smaller `EPAr' model which we use throughout the Analysis section for time and memory feasibility. We also use the last hidden state of the encoding LSTM-RNN to get the compact representation for all supporting documents in case of smaller model, in contrast to self-attention (\secref{sec:selfattention} in Appendix) as in the full-size `EPAr' model. 
The encoding LSTM-RNN ~\citep{hochreiter1997lstm} has 100-d hidden size for our `EPAr' model whereas the smaller version has 20-d hidden size. The embedded GRU~\citep{cho2014gru} and the LSTM in our Evidence Assembler have the hidden dimension of 80.
In practice, we only apply TF-IDF based retrieval procedure to our Document Explorer and Answer Proposer during inference, and during training time we use the full set of supporting documents as the input. This is because we observed that the Document Explorer overfits faster in the reduced document-search space.
For the Evidence Assembler, we employ both the TF-IDF retrieval and Document Explorer to get the `reasoning tree' of documents, at both training and testing time.
We refer to the \secref{app:impl-details} in the appendix for the implementation details of our MedHop models.

\subsection{Results}
We first evaluate our system on the WikiHop dataset.
For a fair comparison to recent works~\cite{Cao2018question,song2018exploring,raison2018weaver}, we report our ``EPAr'' with 300-d embeddings and 100-d hidden size of the encoding LSTM-RNN. 
As shown in \tabref{table:main}, EPAr achieves 67.2\% accuracy on the dev set, outperforming all published models, and achieves 69.1\% accuracy on the hidden test set, which is competitive with the current state-of-the-art result.\footnote{Note that there also exists a recent anonymous unpublished entry on the leaderboard with 70.9\% accuracy, which is concurrent to our work. Also note that our system achieves these strong accuracies even without using pretrained language model representations like ELMo ~\citep{Peters2018DeepCW} or BERT ~\citep{devlin2018bert}, which have been known to give significant improvements in machine comprehension and QA tasks. We leave these gains for future work.}

\begin{table}[t]
\centering
\begin{small}
\begin{tabular}[t]{lcc}
\toprule
& Dev & Test\\
\midrule
BiDAF~\cite{welbl2017qangaroo}$^\star$ & - & 42.9\\
Coref-GRU~\cite{dhingra2018corefgru} & 56.0 & 59.3\\
WEAVER~\cite{raison2018weaver} & 64.1 & 65.3\\
MHQA-GRN~\cite{song2018exploring} & 62.8 & 65.4 \\
Entity-GCN~\cite{Cao2018question} & 64.8 & 67.6 \\
BAG~\cite{Cao2019bag} & 66.5 & 69.0 \\
CFC~\cite{zhong2019coarse} & 66.4 & 70.6 \\
\midrule
EPAr (Ours) & \textbf{67.2} & 69.1\\
\bottomrule
\end{tabular}
\vspace{-5pt}
\caption{Dev set and Test set accuracy on \textsc{WikiHop} dataset. The model marked with $\star$ does not use candidates and directly predict the answer span. EPAr is our system with TF-IDF retrieval, Document Explorer, Answer Proposer and Evidence Assembler.
\vspace{-3pt}
}
\label{table:main}
\end{small}
\end{table}

\begin{table}[t]
\centering
\begin{small}
\begin{tabular}[t]{lccc}
\toprule
\multirow{2}*{}  & follow & follow & \multirow{2}*{full}\\
& + multiple & + single & \\
\midrule
BiDAF Baseline & 62.8 & 63.1 & 58.4 \\
DE+AP+EA$^\star$ & 65.2 & 66.9 & 61.1\\
AP+EA & 68.7 & 67.0 & 62.8\\
DE+AP+EA & 69.4 & 70.6 & 64.7\\
DE+AP+EA$^\dagger$ & 71.8 & \textbf{73.8} & 66.9  \\
DE+AP+EA$^\dagger$+SelfAttn & \textbf{73.5} & 72.9 & \textbf{67.2}  \\
\bottomrule
\end{tabular}
\vspace{-5pt}
\caption{Ablation accuracy on \textsc{WikiHop} dev set. The model marked with $^\star$ does not use the TFIDF-based document retrieval procedure.
The models marked with $^\dagger$ are our full EPAr systems with 300-d word embeddings and 100-d LSTM-RNN hidden size (same as the last row of \tabref{table:main}), while the 4th row represents the smaller EPAr system.
\vspace{-8pt}
}
\label{table:multiple-single}
\end{small}
\end{table}

\begin{figure}[t]
\centering
\includegraphics[width=0.45\textwidth]{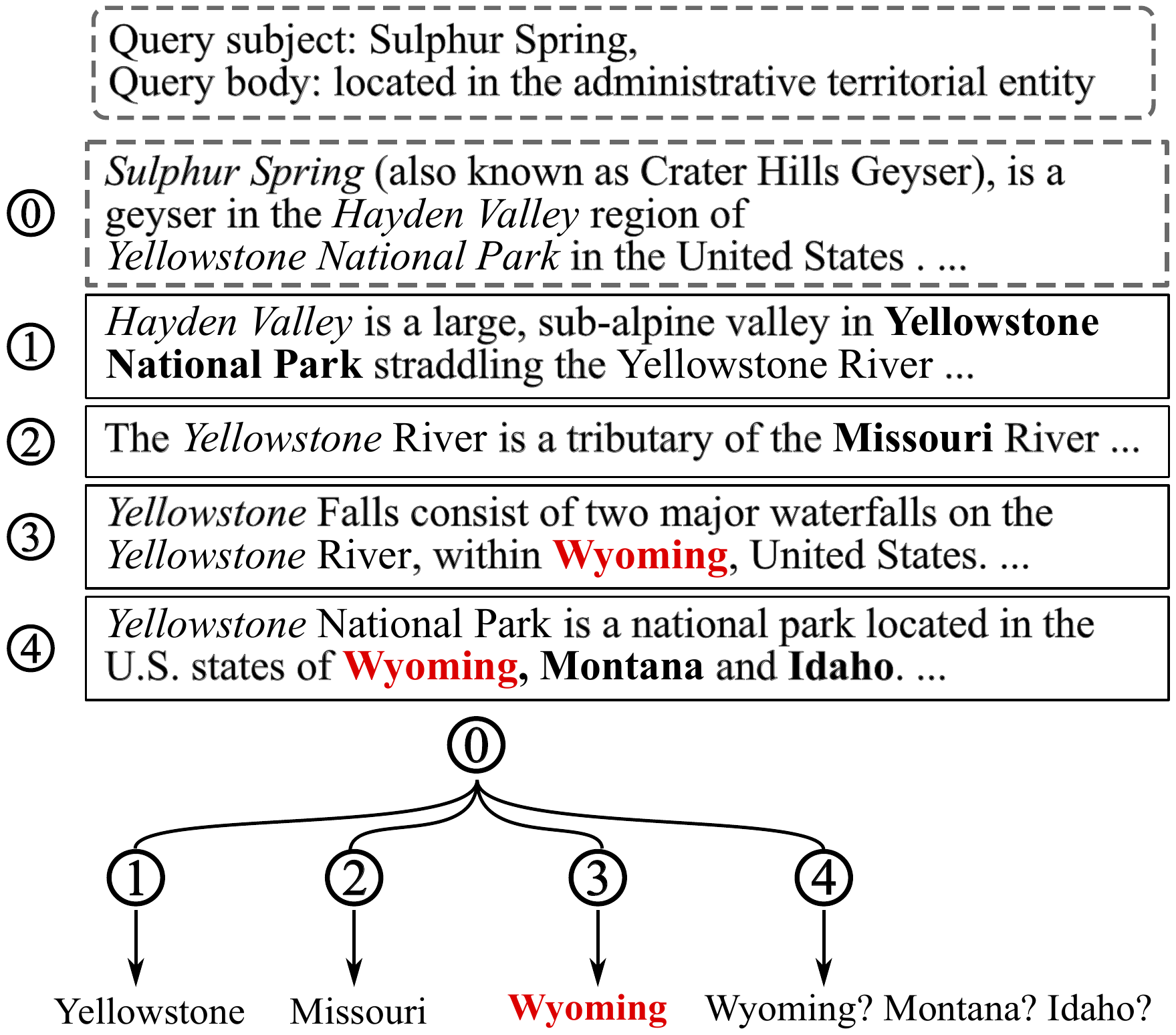}
\vspace{-3pt}
\caption{A `reasoning tree' with 4 leaves that lead to different answers (marked in bold). The ground-truth answer is marked in red additionally.}
\label{fig:tree-example-v1}
\vspace{-8pt}
\end{figure}

Next, in \tabref{table:multiple-single}, we further evaluate our EPAr system (and its smaller-sized and ablated versions) on the ``follows + multiple", ``follows + single", and the full development set.
First, note that on the full development set, our smaller system (``DE+AP+EA'') achieves statistically significant (p-value \(< 0.01\))\footnote{All stat. signif. is based on bootstrapped randomization test with 100K samples~\citep{Efron:94}.} improvements over the BiDAF baseline and is also comparable to~\citet{Cao2018question} on the development set (64.7 vs. 64.8).\footnote{For time and memory feasibility, we use this smaller strong model with 100-d word embeddings and 20-d LSTM-RNN hidden size (similar to baselines in~\newcite{welbl2017qangaroo}) in all our analysis/ablation results (including \secref{sec:analysis}).} 
Moreover, we see that EPAr is able to achieve high accuracy in both the examples that require multi-hop reasoning (``follows + multiple"), and other cases where a single document suffices for correctly answering the question (``follows + single"), suggesting that our system is able to adjust to examples of different reasoning requirements.
The evaluation results further demonstrate that our Document Explorer combined with TF-IDF-based retrieval (row `DE+AP+EA') consistently outperforms TF-IDF alone (row `AP+EA') or the Document Explorer without TF-IDF (row `DE+AP+EA$^\star$' in \tabref{table:multiple-single}), showing that our 2-hop TF-IDF document retrieval procedure is able to broadly identify relevant documents and further aid our Document Explorer by reducing its search space.
Finally, comparing the last two rows in \tabref{table:multiple-single} shows that using self-attention~\citep{zhong2019coarse} to compute the document representation can further improve the full-sized system.
We show an example of the `reasoning tree' constructed by the Document Explorer and the correct answer predicted by the Evidence Assembler in \figref{fig:tree-example-v1}.

\begin{table}[t]
\centering
\begin{small}
\begin{tabular}[t]{lcc}
\toprule
\multirow{2}*{} & Test & \multirow{2}*{Test} \\
& (Masked) &\\
\midrule
FastQA$^\star$~\cite{Weissenborn2017MakingNQ} & 23.1 & 31.3\\
BiDAF$^\star$~\cite{seo2016bidaf} & 33.7 & 47.8 \\
CoAttention  & - & 58.1 \\
Most Frequent Candidate$^\star$ & 10.4 & 58.4 \\
\midrule
EPAr (Ours) & \textbf{41.6} & \textbf{60.3}\\
\bottomrule
\end{tabular}
\vspace{-3pt}
\caption{Test set accuracy on \textsc{MedHop} dataset. The results marked with $^\star$ are reported in \cite{welbl2017qangaroo}.
}

\label{table:medhop_main}
\vspace{-10pt}
\end{small}
\end{table}

We report our system's accuracy on the MedHop dataset in \tabref{table:medhop_main}. Our best system achieves 60.3 on the hidden test set\footnote{The masked MedHop test set results use the smaller size model, because this performed better on the masked dev set.}, outperforming all current models on the leaderboard.
However, as reported by \newcite{welbl2017qangaroo}, the original MedHop dataset suffers from a candidate frequency imbalance issue that can be exploited by certain heuristics like the `Most Frequent Candidate' in \tabref{table:medhop_main}. To eliminate this bias and to test our system's ability to conduct multi-hop reasoning using the context, we additionally evaluate our system on the masked version of MedHop, where every candidate expression is replaced randomly using 100 unique placeholder tokens so that models can only rely on the context to comprehend every candidate. 
Our model achieves 41.6\% accuracy in this ``masked" setting, outperforming all previously published works by a large margin. 

\section{Analysis}
\label{sec:analysis}
In this section, we present a series of new analyses and comparisons in order to understand the contribution from each of our three modules and demonstrate their advantages over other corresponding baselines and heuristics.

\subsection{Reasoning Chain Recovery Tests}

\begin{table}[t]
\centering
\begin{small}
\begin{tabular}[t]{lccccc}
\toprule
& R@1 & R@2 & R@3 & R@4 & R@5\\
\midrule
Random & 11.2 & 17.3 & 27.6 & 40.8 & 50.0\\
1-hop TFIDF & 32.7 & 48.0 & 56.1 & 63.3 & 70.4\\
2-hop TFIDF & 42.9 & 56.1 & 70.4 & 78.6 & 82.7\\
DE  & 38.8 & 50.0 & 65.3 & 73.5 & 83.7\\
TFIDF+DE & \textbf{44.9} & \textbf{64.3} & \textbf{77.6} & \textbf{82.7} & \textbf{90.8}\\
\bottomrule
\end{tabular}
\vspace{-5pt}
\caption{Recall-k score is the \% of examples where one of the human-annotated reasoning chains is recovered in the top-k root-to-leaf paths in the `reasoning tree'. `TFIDF+DE' is the combination of the 2-hop TF-IDF retrieval procedure and our Document Explorer.}
\label{table:analysis_golden_chain}
\vspace{-3pt}
\end{small}
\end{table}

\begin{table}[t]
\centering
\begin{small}
\begin{tabular}[t]{lccccc}
\toprule
& R@1 & R@2 & R@3 & R@4 & R@5\\
\midrule
Random & 39.9 & 51.4 & 60.2 & 67.8 & 73.5\\
1-hop TFIDF & 38.4 & 48.5 & 58.6 & 67.4 & 73.7 \\
2-hop TFIDF & 38.4 & 58.7 & 70.2 & 77.2 & 81.6\\
DE & \textbf{52.5} & \textbf{70.2} & \textbf{80.3} & \textbf{85.8} & \textbf{89.0}\\
TFIDF+DE & 52.2 & 69.0 & 77.8 & 82.2 & 85.2\\
\bottomrule
\end{tabular}
\vspace{-5pt}
\caption{Recall-k score is the percentage of examples where the ground-truth answer is present in the top-k root-to-leaf path in the `reasoning tree'. `TFIDF+DE' is the combination of the 2-hop TFIDF retrieval procedure and our Document Explorer.}
\vspace{-10pt}
\label{table:analysis_de}
\end{small}
\end{table}

We compare our Document Explorer with two TF-IDF-based document selectors for their ability to recover the reasoning chain of documents.
The 1-hop TF-IDF selector selects the top $k+1$ documents with the highest TF-IDF score w.r.t. the query subject.
The 2-hop TF-IDF selector, as in \secref{ssec:ret_and_enc}, first selects the top-1 TF-IDF document w.r.t. the query subject and then selects the top $k$ remaining documents based on the TF-IDF score with respect to the first selected document. Finally, we also compare to our final combination of 2-hop TF-IDF and  Document Explorer.

\paragraph{Human Evaluation:}
We collect human-annotated reasoning chains for 100 documents from the ``follows + multiple" dev set, and compare these to the `reasoning tree' constructed by our Document Explorer to assess its ability to discover the hidden reasoning chain from the entire pool of supporting documents.
For each example, human annotators (external, English-speaking) select two of the smallest set of documents, from which they can reason to find the correct answer from the question.
As shown in \tabref{table:analysis_golden_chain}, our Document Explorer combined with 2-hop TF-IDF (row `TFIDF+DE') obtains higher golden-chain recall scores compared to the two TFIDF-based document retrieval heuristics (row `1-hop TFIDF' and `2-hop TFIDF') alone or the Document Explorer without TF-IDF (row `DE').

\paragraph{Answer Span Test:}
We also test our Document Explorer's ability to find the document with mentions of the ground-truth answer.
Logically, the fact that the answer appears in one of the documents in the `reasoning tree' signals higher probability that our modules at the following stages could predict the correct answer.
As shown in \tabref{table:analysis_de}, our Document Explorer receives significantly higher answer-span recall scores compared to the two TF-IDF-based document selectors.\footnote{In this test, the Document Explorer alone outperforms its combination with the 2-hop TF-IDF retrieval. In practice, our system employs both procedures due to the advantage shown in both empirical results (\tabref{table:multiple-single}) and analysis (\tabref{table:analysis_golden_chain}).
}

\begin{table}[t]
\centering
\begin{small}
\begin{tabular}[t]{lccc}
\toprule
\multirow{2}*{}& \multirow{2}*{full} & follows & follows\\
& & + multiple & + single \\
\midrule
Full-doc & 63.1 & 68.4 & 69.0\\
Lead-1 & 63.6 & 68.7 & 70.2\\
AP w.o. attn & 63.3 & 68.3& 69.6 \\
AP & \textbf{64.7} & \textbf{69.4} & \textbf{70.6}\\
\bottomrule
\end{tabular}
\vspace{-5pt}
\caption{Answer Proposer comparison study. ``Follows + multiple" and ``follows + single" are the subsets of dev set as described in \secref{ssec:dataset}.
}
\label{table:analysis_ap}
\end{small}
\end{table}

\subsection{Answer Proposer Comparisons}
We compare our Answer Proposer with two rule-based sentence extraction heuristics for the ability to extract salient information from every reasoning chain. For most documents in the WikiHop dataset, the first sentence is comprised of the most salient information from that document. Hence, we construct one baseline that concatenates the first sentence from each selected document as the input to the Evidence Assembler. 
We also show results of combining all the full documents as the synthesized context instead of selecting one sentence from every document. 
We further present a lighter neural-model baseline that directly proposes the answer from the leaf document without first creating its ancestor-aware representation.
As shown in \tabref{table:analysis_ap}, the system using sentences selected by our Answer Proposer outperforms both rule-based heuristics (row 1 and 2) and the simple neural baseline (row 3).

\begin{table}[t]
\centering
\begin{small}
\begin{tabular}[t]{lccc}
\toprule
\multirow{2}*{}& \multirow{2}*{full} & follows & follows\\
& & + multiple & + single \\
\midrule
    Single-chain & 59.9 & 64.3 & 63.8 \\
    Avg-vote & 54.6 & 56.3 & 55.6 \\
    Max-vote & 51.5 & 53.9 & 53.3 \\
    w. Reranker & 60.6 & 65.1 & 65.5 \\
    w. Assembler & \textbf{64.7} & \textbf{69.4} & \textbf{70.6}\\
\bottomrule
\end{tabular}
\caption{Evidence Assembler comparison study: Reranker (described in the appendix) rescores the documents selected by the Document Explorer.}
    \label{table:analysis_ea}
    \vspace{-5pt}
\end{small}
\end{table}
\subsection{Assembler Ablations}
\label{ssec:assembler_analysis}
In order to justify our choice of building an Assembler, we build a 2-module system without the Evidence-Assembler stage by applying the Answer Proposer to only the top-1 reasoning chain in the tree. We also present two voting heuristics that selects the final answer by taking the average/maximum prediction probability from the Answer Proposer on all document chains.
Furthermore, we compare our Evidence Assembler with an alternative model that, instead of assembling information from all reasoning chains, reranks all chains and their proposed answers to select the top-1 answer prediction.
As shown in \tabref{table:analysis_ea}, the full system with the Assembler achieves significant improvements over the 2-module system. This demonstrates the importance of the Assembler in enabling information aggregation over multiple reasoning chains.
The results further show that our Assembler is better than the reranking alternative.
\begin{figure*}[t]
\centering
\includegraphics[width=0.98\textwidth]{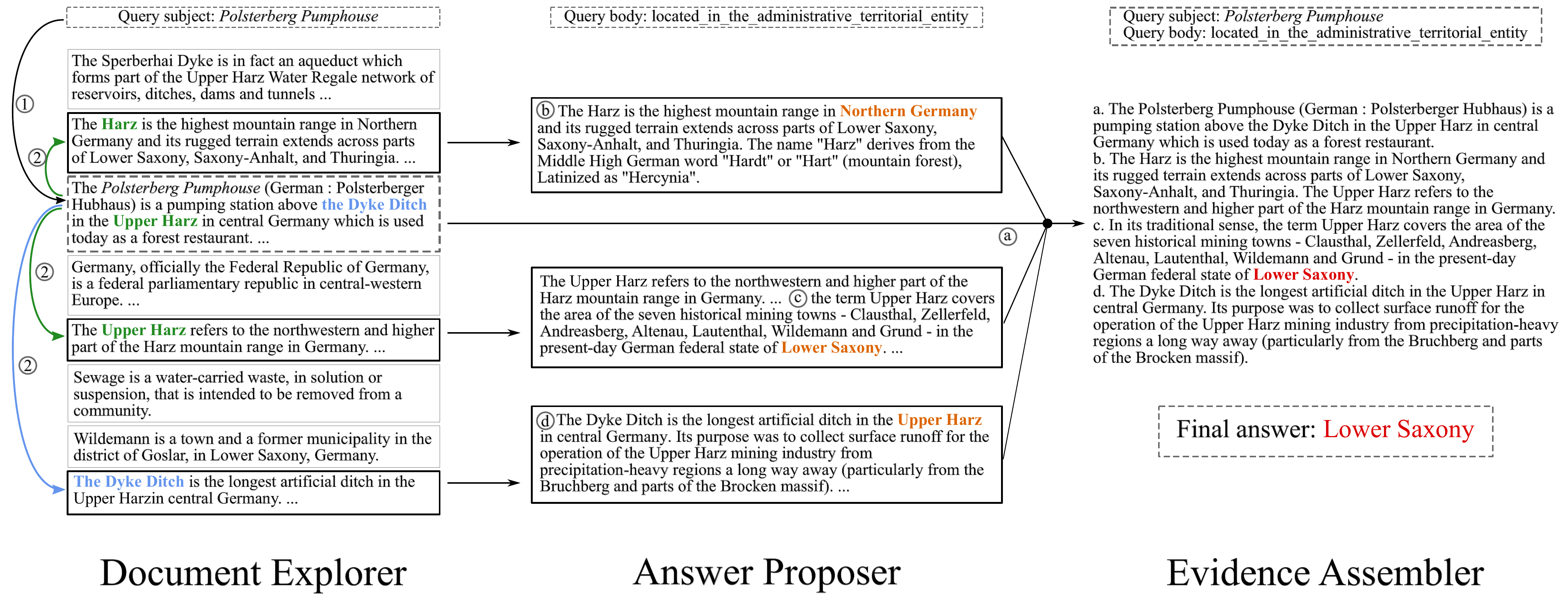}
\caption{An example of our 3-stage EPAr system exploring relevant documents, proposing candidate answers, and then assembling extracted evidence to make the final prediction. 
\label{fig:architecture}
}
\end{figure*}

\subsection{Multi-hop Reasoning Example}
We visualize the 3-stage reasoning procedure of our EPAr system in \figref{fig:architecture}.
As shown in the left of \figref{fig:architecture}, the Document Explorer first locates the root document (``The Polsterberg Pumphouse ...") based on the query subject. It then finds three more documents that are related to the root document, constructing three document chains. The Answer Proposer proposes a candidate answer from each of the three chains selected by the Document Explorer.
Finally, the Evidence Assembler selects key sentences from all documents in the constructed document chains and makes the final prediction (``Lower Saxony").
\section{Related Works}

The last few years have witnessed significant progress on text-based machine reading comprehension and question answering (MRC-QA) including cloze-style blank-filling tasks~\citep{Hermann:cnndm}, open-domain QA~\citep{yang2015wikiqa}, answer span prediction~\citep{rajpurkar2016squad,rajpurkar2018squad2}, and generative QA~\citep{nguyen2016msmarco}.
However, all of the above datasets are confined to a single-document context per question setup. \citet{joshi2017triviaqa} extended the task to the multi-document regime, with some examples requiring cross-sentence inference.
Earlier attempts in multi-hop MRC focused on reasoning about the relations in a knowledge base~\citep{Jain2016Question,zhou2018multirelation,lin2018multi} or tables \citep{yin2015neural}.
QAngaroo WikiHop and MedHop~\citep{welbl2017qangaroo}, on the other hand, are created as \emph{natural language} MRC tasks. 
They are designed in a way such that the evidence required to answer a query could be spread across multiple documents.
Thus, finding some evidence requires building a reasoning chain from the query with intermediate inference steps, which poses extra difficulty for MRC-QA systems.
HotpotQA~\cite{yang2018hotpotqa} is another recent multi-hop dataset which focuses on four different reasoning paradigms.

The emergence of large-scale MRC datasets has led to innovative neural models such as co-attention~\citep{xiong2016dynamic}, bi-directional attention flow~\citep{seo2016bidaf}, and gated attention~\citep{dhingra2017GA}, all of which are meticulously designed to solve single-document MRC tasks.
\citet{clark2017simple} and \citet{chen2017drqa} used a simple TF-IDF based document-selection procedure to find the context that is most relevant to the query for multi-document QA. 
However, this 1-hop, similarity-based selection process would fail on multi-hop reading-comprehension datasets like WikiHop because the query subject and the answer could appear in different documents.
On the other hand, our Document Explorer can discover the document with the answer ``Loon op Zand" (in \figref{fig:example-intro-a}) by iteratively selecting relevant documents and encoding the hinge words ``Efteling" and ``Kaatsheuvel" in its memory.

Recently, \citet{dhingra2018corefgru} leveraged coreference annotations from an external system to connect the entities.
\citet{song2018exploring} and \citet{Cao2018question} utilized Graph Convolutional Networks~\citep{Kipf2017graphconv} and Graph Recurrent Networks~\citep{song2018graph, zhang2018slstm} to model the relations between entities.
Recently, \citet{Cao2019bag} extended the Graph Convolutional Network in \citet{Cao2018question} by introducing bi-directional attention between the entity graph and query.
By connecting the entities, these models learn the inference paths for multi-hop reasoning.
Our work differs in that our system learns the relation implicitly without the need of any human-annotated relation. Recently,~\newcite{zhong2019coarse} used hierarchies of co-attention and self-attention to combine evidence from multiple scattered documents.
Our novel 3-module architecture is inspired by previous 2-module selection architectures for MRC~\citep{Choi2017CoarsetoFineQA}.
Similarly, \citet{Wang2018R3RR} first selected relevant content by ranking documents and then extracted the answer span.
\citet{min2018efficient} selected relevant sentences from long documents in a single-document setup and achieved faster speed and robustness against adversarial corruption.
However, none of these models are built for multi-hop MRC where our EPAr system shows great effectiveness.

\section{Conclusion}
We presented an interpretable 3-module, multi-hop, reading-comprehension system `EPAr' which constructs a `reasoning tree', proposes an answer candidate for every root-to-leaf chain, and merges key information from all reasoning chains to make the final prediction.
On WikiHop, our system outperforms all published models on the dev set, and achieves results competitive with the current state-of-the-art on the test set. On MedHop, our system outperforms all previously published models on the leaderboard test set.
We also presented multiple reasoning-chain recovery tests for the explainability of our system's reasoning capabilities.

\section{Acknowledgement}
We would like to thank Johannes Welbl for helping test our system on WikiHop and MedHop. We thank the reviewers for their helpful comments. This work was supported by DARPA (YFA17-D17AP00022), Google Faculty Research
Award, Bloomberg Data Science Research Grant, Salesforce Deep Learning Research Grant, Nvidia GPU awards, Amazon AWS, and Google Cloud Credits. The views contained in this article are those of the authors and not of the funding agency.

\bibliography{main}

\begin{thebibliography}{40}
\expandafter\ifx\csname natexlab\endcsname\relax\def\natexlab#1{#1}\fi

\bibitem[{{Bahdanau} et~al.(2015){Bahdanau}, {Cho}, and {Bengio}}]{Bahdanau:14}
D.~{Bahdanau}, K.~{Cho}, and Y.~{Bengio}. 2015.
\newblock {Neural Machine Translation by Jointly Learning to Align and
  Translate}.
\newblock In \emph{Third International Conference on Learning Representations}.

\bibitem[{Cao et~al.(2019)Cao, Fang, and Tao}]{Cao2019bag}
Yu~Cao, Meng Fang, and Dacheng Tao. 2019.
\newblock {BAG:} bi-directional attention entity graph convolutional network
  for multi-hop reasoning question answering.
\newblock In \emph{NAACL-HLT}.

\bibitem[{Chandar et~al.(2016)Chandar, Ahn, Larochelle, Vincent, Tesauro, and
  Bengio}]{chandar2016hierarchical}
Sarath Chandar, Sungjin Ahn, Hugo Larochelle, Pascal Vincent, Gerald Tesauro,
  and Yoshua Bengio. 2016.
\newblock Hierarchical memory networks.
\newblock \emph{arXiv preprint arXiv:1605.07427}.

\bibitem[{Chen et~al.(2017)Chen, Fisch, Weston, and Bordes}]{chen2017drqa}
Danqi Chen, Adam Fisch, Jason Weston, and Antoine Bordes. 2017.
\newblock Reading {Wikipedia} to answer open-domain questions.
\newblock In \emph{ACL}.

\bibitem[{Cho et~al.(2014)Cho, van Merrienboer, G{\"{u}}l{\c{c}}ehre, Bougares,
  Schwenk, and Bengio}]{cho2014gru}
Kyunghyun Cho, Bart van Merrienboer, {\c{C}}aglar G{\"{u}}l{\c{c}}ehre, Fethi
  Bougares, Holger Schwenk, and Yoshua Bengio. 2014.
\newblock Learning phrase representations using {RNN} encoder-decoder for
  statistical machine translation.
\newblock In \emph{EMNLP}.

\bibitem[{Choi et~al.(2017)Choi, Hewlett, Uszkoreit, Polosukhin, Lacoste, and
  Berant}]{Choi2017CoarsetoFineQA}
Eunsol Choi, Daniel Hewlett, Jakob Uszkoreit, Illia Polosukhin, Alexandre
  Lacoste, and Jonathan Berant. 2017.
\newblock Coarse-to-fine question answering for long documents.
\newblock In \emph{ACL}.

\bibitem[{Clark and Gardner(2018)}]{clark2017simple}
Christopher Clark and Matt Gardner. 2018.
\newblock Simple and effective multi-paragraph reading comprehension.
\newblock In \emph{Proceedings of the 56th Annual Meeting of the Association
  for Computational Linguistics}.

\bibitem[{De~Cao et~al.(2018)De~Cao, Aziz, and Titov}]{Cao2018question}
Nicola De~Cao, Wilker Aziz, and Ivan Titov. 2018.
\newblock Question answering by reasoning across documents with graph
  convolutional networks.
\newblock \emph{arXiv preprint arXiv:1808.09920}.

\bibitem[{Devlin et~al.(2018)Devlin, Chang, Lee, and
  Toutanova}]{devlin2018bert}
Jacob Devlin, Ming-Wei Chang, Kenton Lee, and Kristina Toutanova. 2018.
\newblock Bert: Pre-training of deep bidirectional transformers for language
  understanding.

\bibitem[{Dhingra et~al.(2018)Dhingra, Jin, Yang, Cohen, and
  Salakhutdinov}]{dhingra2018corefgru}
Bhuwan Dhingra, Qiao Jin, Zhilin Yang, William~W Cohen, and Ruslan
  Salakhutdinov. 2018.
\newblock Neural models for reasoning over multiple mentions using coreference.
\newblock In \emph{Proceedings of the 16th Annual Conference of the North
  American Chapter of the Association for Computational Linguistics: Human
  Language Technologies}.

\bibitem[{Dhingra et~al.(2017)Dhingra, Liu, Yang, Cohen, and
  Salakhutdinov}]{dhingra2017GA}
Bhuwan Dhingra, Hanxiao Liu, Zhilin Yang, William Cohen, and Ruslan
  Salakhutdinov. 2017.
\newblock Gated-attention readers for text comprehension.
\newblock In \emph{Proceedings of the 55th Annual Meeting of the Association
  for Computational Linguistics (Volume 1: Long Papers)}, pages 1832--1846,
  Vancouver, Canada. Association for Computational Linguistics.

\bibitem[{Efron and Tibshirani(1994)}]{Efron:94}
Bradley Efron and Robert~J Tibshirani. 1994.
\newblock \emph{An introduction to the bootstrap}.
\newblock CRC press.

\bibitem[{Hermann et~al.(2015)Hermann, Kocisky, Grefenstette, Espeholt, Kay,
  Suleyman, and Blunsom}]{Hermann:cnndm}
Karl~Moritz Hermann, Tomas Kocisky, Edward Grefenstette, Lasse Espeholt, Will
  Kay, Mustafa Suleyman, and Phil Blunsom. 2015.
\newblock Teaching machines to read and comprehend.
\newblock In \emph{Advances in Neural Information Processing Systems}, pages
  1693--1701.

\bibitem[{Hochreiter and Schmidhuber(1997)}]{hochreiter1997lstm}
Sepp Hochreiter and J{\"u}rgen Schmidhuber. 1997.
\newblock Long short-term memory.
\newblock \emph{Neural computation}, 9(8):1735--1780.

\bibitem[{Jain(2016)}]{Jain2016Question}
Sarthak Jain. 2016.
\newblock Question answering over knowledge base using factual memory networks.
\newblock In \emph{Proceedings of the NAACL Student Research Workshop}.
  Association for Computational Linguistics.

\bibitem[{Joshi et~al.(2017)Joshi, Choi, Weld, and
  Zettlemoyer}]{joshi2017triviaqa}
Mandar Joshi, Eunsol Choi, Daniel~S Weld, and Luke Zettlemoyer. 2017.
\newblock Triviaqa: A large scale distantly supervised challenge dataset for
  reading comprehension.
\newblock \emph{arXiv preprint arXiv:1705.03551}.

\bibitem[{Kingma and Ba(2014)}]{Kingma2014AdamAM}
Diederik~P. Kingma and Jimmy Ba. 2014.
\newblock Adam: A method for stochastic optimization.
\newblock \emph{CoRR}.

\bibitem[{Kipf and Welling(2017)}]{Kipf2017graphconv}
Thomas~N. Kipf and Max Welling. 2017.
\newblock Semi-supervised classification with graph convolutional networks.
\newblock In \emph{ICLR}.

\bibitem[{Lin et~al.(2018)Lin, Socher, and Xiong}]{lin2018multi}
Xi~Victoria Lin, Richard Socher, and Caiming Xiong. 2018.
\newblock Multi-hop knowledge graph reasoning with reward shaping.
\newblock In \emph{EMNLP}.

\bibitem[{Min et~al.(2018)Min, Zhong, Socher, and Xiong}]{min2018efficient}
Sewon Min, Victor Zhong, Richard Socher, and Caiming Xiong. 2018.
\newblock Efficient and robust question answering from minimal context over
  documents.
\newblock In \emph{Proceedings of the 56th Annual Meeting of the Association
  for Computational Linguistics (Volume 1: Long Papers)}, pages 1725--1735.
  Association for Computational Linguistics.

\bibitem[{Nguyen et~al.(2016)Nguyen, Rosenberg, Song, Gao, Tiwary, Majumder,
  and Deng}]{nguyen2016msmarco}
Tri Nguyen, Mir Rosenberg, Xia Song, Jianfeng Gao, Saurabh Tiwary, Rangan
  Majumder, and Li~Deng. 2016.
\newblock Ms marco: A human generated machine reading comprehension dataset.
\newblock \emph{arXiv preprint arXiv:1611.09268}.

\bibitem[{Pennington et~al.(2014)Pennington, Socher, and
  Manning}]{pennington2014glove}
Jeffrey Pennington, Richard Socher, and Christopher~D Manning. 2014.
\newblock Glove: Global vectors for word representation.
\newblock In \emph{Conference on Empirical Methods in Natural Language
  Processing (EMNLP)}.

\bibitem[{Peters et~al.(2018)Peters, Neumann, Iyyer, Gardner, Clark, Lee, and
  Zettlemoyer}]{Peters2018DeepCW}
Matthew~E. Peters, Mark Neumann, Mohit Iyyer, Matt Gardner, Christopher Clark,
  Kenton Lee, and Luke~S. Zettlemoyer. 2018.
\newblock Deep contextualized word representations.
\newblock In \emph{NAACL-HLT}.

\bibitem[{Raison et~al.(2018)Raison, Mazar{\'e}, Das, and
  Bordes}]{raison2018weaver}
Martin Raison, Pierre-Emmanuel Mazar{\'e}, Rajarshi Das, and Antoine Bordes.
  2018.
\newblock Weaver: Deep co-encoding of questions and documents for machine
  reading.
\newblock \emph{arXiv preprint arXiv:1804.10490}.

\bibitem[{Rajpurkar et~al.(2018)Rajpurkar, Jia, and
  Liang}]{rajpurkar2018squad2}
P.~Rajpurkar, R.~Jia, and P.~Liang. 2018.
\newblock Know what you don't know: Unanswerable questions for {SQuAD}.
\newblock In \emph{Association for Computational Linguistics (ACL)}.

\bibitem[{Rajpurkar et~al.(2016)Rajpurkar, Zhang, Lopyrev, and
  Liang}]{rajpurkar2016squad}
Pranav Rajpurkar, Jian Zhang, Konstantin Lopyrev, and Percy Liang. 2016.
\newblock Squad: 100,000+ questions for machine comprehension of text.
\newblock In \emph{Conference on Empirical Methods in Natural Language
  Processing (EMNLP)}.

\bibitem[{Seo et~al.(2017)Seo, Kembhavi, Farhadi, and
  Hajishirzi}]{seo2016bidaf}
Minjoon Seo, Aniruddha Kembhavi, Ali Farhadi, and Hannaneh Hajishirzi. 2017.
\newblock Bidirectional attention flow for machine comprehension.
\newblock In \emph{International Conference on Learning Representations
  (ICLR)}.

\bibitem[{Song et~al.(2018{\natexlab{a}})Song, Wang, Yu, Zhang, Florian, and
  Gildea}]{song2018exploring}
Linfeng Song, Zhiguo Wang, Mo~Yu, Yue Zhang, Radu Florian, and Daniel Gildea.
  2018{\natexlab{a}}.
\newblock Exploring graph-structured passage representation for multi-hop
  reading comprehension with graph neural networks.
\newblock \emph{arXiv preprint arXiv:1809.02040}.

\bibitem[{Song et~al.(2018{\natexlab{b}})Song, Zhang, Wang, and
  Gildea}]{song2018graph}
Linfeng Song, Yue Zhang, Zhiguo Wang, and Daniel Gildea. 2018{\natexlab{b}}.
\newblock A graph-to-sequence model for amr-to-text generation.
\newblock In \emph{Proceedings of the 56th Annual Meeting of the Association
  for Computational Linguistics (Volume 1: Long Papers)}, pages 1616--1626.
  Association for Computational Linguistics.

\bibitem[{Srivastava et~al.(2015)Srivastava, Greff, and
  Schmidhuber}]{srivastava2015highway}
Rupesh~Kumar Srivastava, Klaus Greff, and J{\"{u}}rgen Schmidhuber. 2015.
\newblock Highway networks.
\newblock In \emph{International Conference on Machine Learning (ICML)}.

\bibitem[{Wang et~al.(2018)Wang, Yu, Guo, Wang, Klinger, Zhang, Chang, Tesauro,
  Zhou, and Jiang}]{Wang2018R3RR}
Shuohang Wang, Mo~Yu, Xiaoxiao Guo, Zhiguo Wang, Tim Klinger, Wei Zhang, Shiyu
  Chang, Gerald Tesauro, Bowen Zhou, and Jing Jiang. 2018.
\newblock R3: Reinforced ranker-reader for open-domain question answering.
\newblock In \emph{AAAI}.

\bibitem[{Weissenborn et~al.(2017)Weissenborn, Wiese, and
  Seiffe}]{Weissenborn2017MakingNQ}
Dirk Weissenborn, Georg Wiese, and Laura Seiffe. 2017.
\newblock Making neural qa as simple as possible but not simpler.
\newblock In \emph{CoNLL}.

\bibitem[{Welbl et~al.(2017)Welbl, Stenetorp, and Riedel}]{welbl2017qangaroo}
Johannes Welbl, Pontus Stenetorp, and Sebastian Riedel. 2017.
\newblock Constructing datasets for multi-hop reading comprehension across
  documents.
\newblock In \emph{TACL}.

\bibitem[{Xiong et~al.(2017)Xiong, Zhong, and Socher}]{xiong2016dynamic}
Caiming Xiong, Victor Zhong, and Richard Socher. 2017.
\newblock Dynamic coattention networks for question answering.
\newblock In \emph{ICLR}.

\bibitem[{Yang et~al.(2015)Yang, Yih, and Meek}]{yang2015wikiqa}
Yi~Yang, Wen-tau Yih, and Christopher Meek. 2015.
\newblock Wikiqa: A challenge dataset for open-domain question answering.
\newblock In \emph{Proceedings of the 2015 Conference on Empirical Methods in
  Natural Language Processing}, pages 2013--2018.

\bibitem[{Yang et~al.(2018)Yang, Qi, Zhang, Bengio, Cohen, Salakhutdinov, and
  Manning}]{yang2018hotpotqa}
Zhilin Yang, Peng Qi, Saizheng Zhang, Yoshua Bengio, William~W Cohen, Ruslan
  Salakhutdinov, and Christopher~D Manning. 2018.
\newblock Hotpotqa: A dataset for diverse, explainable multi-hop question
  answering.
\newblock In \emph{Conference on Empirical Methods in Natural Language
  Processing (EMNLP)}.

\bibitem[{Yin et~al.(2015)Yin, Lu, Li, and Kao}]{yin2015neural}
Pengcheng Yin, Zhengdong Lu, Hang Li, and Ben Kao. 2015.
\newblock Neural enquirer: Learning to query tables.
\newblock \emph{arXiv preprint}.

\bibitem[{Zhang et~al.(2018)Zhang, Liu, and Song}]{zhang2018slstm}
Yue Zhang, Qi~Liu, and Linfeng Song. 2018.
\newblock Sentence-state lstm for text representation.
\newblock In \emph{Proceedings of the 56th Annual Meeting of the Association
  for Computational Linguistics (ACL)}.

\bibitem[{Zhong et~al.(2019)Zhong, Xiong, Keskar, and Socher}]{zhong2019coarse}
Victor Zhong, Caiming Xiong, Nitish~Shirish Keskar, and Richard Socher. 2019.
\newblock Coarse-grain fine-grain coattention network for multi-evidence
  question answering.
\newblock In \emph{ICLR}.

\bibitem[{Zhou et~al.(2018)Zhou, Huang, and Zhu}]{zhou2018multirelation}
Mantong Zhou, Minlie Huang, and Xiaoyan Zhu. 2018.
\newblock An interpretable reasoning network for multi-relation question
  answering.
\newblock In \emph{Proceedings of the 27th International Conference on
  Computational Linguistics}.

\end{thebibliography}
\bibliographystyle{acl_natbib}

\appendix

\section*{Appendix}

\section{Reranker}
We explore an alternative to Evidence Assembler (EA), where instead of selecting key sentences from every root-to-leaf path in the reasoning tree, we use a reranker to rescore the selected documents. Specifically, given a document reasoning-tree of \(t_w\) reasoning chains, we use bidirectional attention \citep{seo2016bidaf} between the last documents in each chain and all the documents from the previous hops in that chain to obtain \(\{\hat{h}_1, \cdots, \hat{h}_{t_{w}}\}\) which are the refined representations of the leaf documents. We then obtain a fixed length document representation as the weighted average of word representations for each of the \(t_w\) documents using similarity with query subject and query body as the weights using function \(\boldsymbol{\alpha}\). We obtain the scores for each of the documents by computing similarity with the answer which that reasoning chain proposes using \(\boldsymbol{\beta}\). (See \secref{ssec:simfunc} below for details of the similarity functions \(\boldsymbol{\alpha}\) and \(\boldsymbol{\beta}\).) 

\section{Self-Attention}
\label{sec:selfattention}
We use self-attention from \citet{zhong2019coarse} to get the compact representation for all supporting documents. Given contextual word representations for the supporting documents \(\mathbf{H} = \{h_1, h_2, \cdots, h_{N^{\prime}}\}\) such that \(h_i \in \mathbb{R}^{K \times 2v}\), we define $\mathrm{Selfattn} (h_i) \rightarrow p_i \in \mathbb{R}^{2v}$ as:
\begin{equation}
\begin{split}
    a_{ik} &= \mathrm{tanh}(W_2 \mathrm{tanh} (W_1 h_i^k + b1) + b2) \\
    \hat{a}_i &= \mathrm{softmax} (a_i) \\
    p_i &= \sum_{k=1}^{K} \hat{a}_{ik} h_i^k
\end{split}
\end{equation}
such that $p_i$ provides the summary of the $i$th document with a vector representation.

\section{Similarity Functions}
\label{ssec:simfunc}

When constructing our 3-module system, we use similarity functions \(\boldsymbol{\alpha}\) and \(\boldsymbol{\beta}\).
The function \(\boldsymbol{\beta}\) is defined as:
\begin{equation}
    \boldsymbol{\beta}(h,c) = \mathbf{W_{\boldsymbol{\beta}_1}}\mathrm{relu}(\mathbf{W_{{\boldsymbol{\beta}}_2}}[h;u;h \circ u] + \mathbf{b_{{\boldsymbol{\beta}}_2}}) + \mathbf{b_{\boldsymbol{{\beta}}_1}} 
\end{equation}
where \(\mathrm{relu}(x) = \mathrm{max}(0,x)\), and \(\circ\) represents element-wise multiplication.
And the function \(\boldsymbol{\alpha}\) is defined as:
\begin{equation}
\begin{aligned}
  \boldsymbol{\alpha}(h,u) =  \mathbf{W_{\boldsymbol{\alpha_2}}}^T( (\mathbf{W_{\boldsymbol{\alpha_1}}}h +  \mathbf{b_{\boldsymbol{\alpha_1}}}) \circ u)\\
  \end{aligned}
\end{equation}
where all trainable weights are marked in bold.

\section{Datasets and Metrics}
We evaluate our 3-module system on QAngaroo \citep{welbl2017qangaroo}, which is a set of two multi-hop reading comprehension datasets: WikiHop and MedHop. WikiHop contains 51K instances, including 44K for training, 5K for development and 2.5K for held out testing. MedHop is a smaller dataset based on the domain of molecular biology. It consists of 1.6K instances for training, 342 for development, and 546 for held out testing. Each instance consists of a query (which can be separated as a query subject and a query body), a set of supporting documents and a list of candidate answers.
For the WikiHop development set, each instance is also annotated as ``follows" or ``not follows", which signifies whether the answer can be inferred from the given set of supporting documents, and ``multiple" or ``single", which tells whether the complete reasoning chain comprises of multiple documents or just a single one. We measure our system's performance on these subsets of the development set that are annotated as ``follows and multiple" and ``follows and single".
This allows us to evaluate our systems on a less noisy version of development set and to investigate their strength in queries requiring different levels of multi-hop reasoning behavior.

\section{Implementation Details}
\label{app:impl-details}
For Medhop, considering the small size of the dataset, we use 20-d hidden size of the encoding LSTM-RNN and the last hidden state of the encoding LSTM-RNN to get compact representation of the documents. We also use a hidden size of 20 for the embedded GRU cell and LSTM in our Evidence Assembler. In addition to that, since ~\citet{welbl2017qangaroo} show the poor performance of TF-IDF model we drop the TF-IDF document retrieval procedure and supervision at the first hop of the Document Explorer (with the document having highest TF-IDF score to query subject).
We train all modules of our system jointly using Adam Optimizer ~\citep{Kingma2014AdamAM} with an initial learning rate of 0.001 and a batch size of 10. We also use a dropout rate of 0.2 in all our linear projection layers, encoding LSTM-RNN and character CNNs.

\end{document}